\documentclass[acmsmall,screen]{acmart}

\usepackage[ruled,vlined]{algorithm2e}
\usepackage{epstopdf}
\usepackage{url}

\AtBeginDocument{%
\providecommand\BibTeX{{%
\normalfont B\kern-0.5em{\scshape i\kern-0.25em b}\kern-0.8em\TeX}}}

\setcopyright{acmcopyright}
\copyrightyear{2021}
\acmYear{2021}
\acmDOI{10.1145/3472722}

\acmJournal{JACM}
\acmVolume{18}
\acmNumber{2}
\acmArticle{46}
\acmMonth{5}

\begin{document}

\title{Learning from Temporal Spatial Cubism for 
Cross-Dataset Skeleton-based Action Recognition}

\author{Yansong Tang}\authornotemark[1]
\email{tys15@tsinghua.org.cn}
\author{Xingyu Liu}
\authornotemark[1]
\author{Xumin Yu}
\author{Danyang Zhang}
\author{Jiwen Lu}
\email{lujiwen@tsinghua.edu.cn}
\author{Jie Zhou}
\authornote{Yansong Tang and Xingyu Liu contributed equally to this research.}
\affiliation{%
  \institution{Beijing National Research Center for Information
Science and Technology, Department of Automation, Tsinghua University}
  \country{China}
}

\renewcommand{\shortauthors}{Tang and Liu, et al.}

\newcommand{\topcaption}{%
\setlength{\abovecaptionskip}{0pt}%
\setlength{\belowcaptionskip}{10pt}%
\caption}
\begin{abstract}
Rapid progress and superior performance have been achieved for skeleton-based action recognition recently.
In this paper, we investigate this problem under a cross-dataset setting, which is a new, pragmatic and challenging task in real-world scenario.
Following the unsupervised domain adaptation (UDA) paradigm,
the action labels are only available on a source dataset, but unavailable on a target dataset in the training stage.
Different from the conventional adversarial learning based approaches for UDA,
we utilize a self-supervision scheme to reduce the domain shift between two skeleton-based action datasets.
Our inspiration is drawn from Cubism, an art genre from the early 20th century, which breaks and reassembles the objects to convey a greater context.
By segmenting and permuting temporal segments or human body parts, we design two self-supervised learning classification tasks to explore
the temporal and spatial dependency of a skeleton-based action and improve the generalization ability of the model.
We conduct experiments on six datasets for skeleton-based action recognition, including three large-scale datasets (NTU RGB+D, PKU-MMD and Kinetics) where new cross-dataset settings and benchmarks are established.
Extensive results demonstrate that our method outperforms state-of-the-art approaches.
The source codes of our model and all the compared methods are available at \url{https://github.com/shanice-l/st-cubism}.
\end{abstract}




\maketitle

\section{Introduction}

\begin{figure}[t]
\centering
\includegraphics[width = 0.7\linewidth]{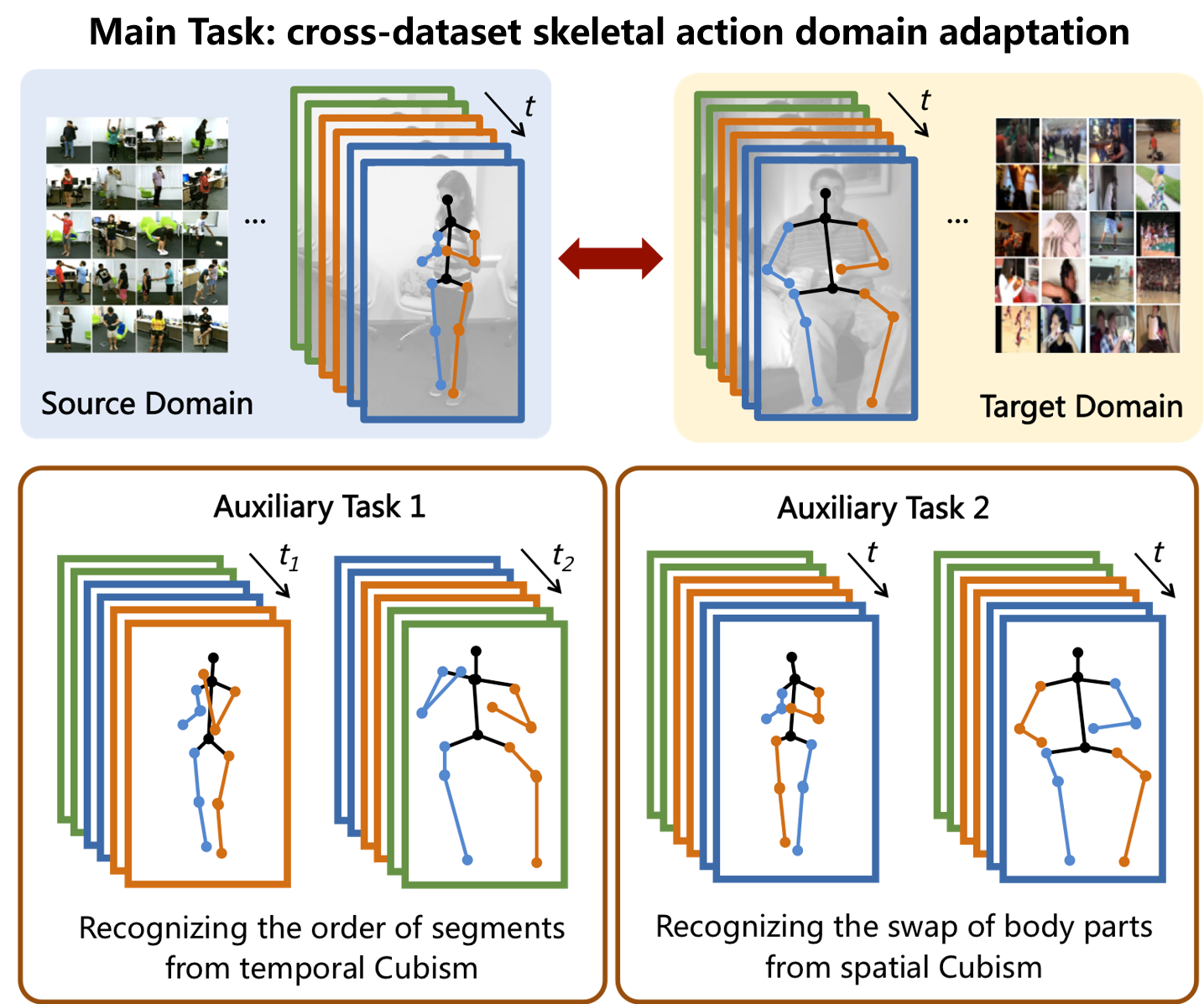} 
\caption{
The goal of this paper is to explore a self-supervised domain adaptation approach of skeleton-based action recognition,
which optimizes a model trained on a source domain (\textit{e.g.,} NTU RGB+D dataset~\cite{DBLP:conf/cvpr/ShahroudyLNW16}) to generalize well on a target domain (\textit{e.g.,} Kinetics dataset~\cite{DBLP:journals/corr/KayCSZHVVGBNSZ17}).
Our main idea is to reduce the domain shift by two auxiliary tasks with self-supervision, which aim to recognize the permutation of the segments in the temporal dimension or in the skeleton joint dimension and guide the network to learn more robust and more general features.
In this figure, the frames in different temporal segments are marked by borders in different colors while the joints assumed as different body parts are also distinguished by their colors.
Fig. \ref{fig:spatial} provides a clearer representation for the spatial Cubism. More detailed description about the auxiliary tasks is referred to sections \ref{subsection:tem_cub} and \ref{subsection:spa_cub}.
All figures are best viewed in color.
}
\label{fig:main}  
\vspace{-0.2cm}
\end{figure}
Skeleton-based action recognition has achieved impressive progress in recent years.
As an evidence, under the cross-view setting of NTU RGB+D dataset~\cite{DBLP:conf/cvpr/ShahroudyLNW16}, the recognition accuracy has been improved from 70.27\% to 96.10\%~\cite{Shi_DGNN_2019_CVPR} significantly.
However, most of the methods apply a fully-supervised learning paradigm where the training and testing data are from the same domain.
Meanwhile, there is a lack of exploration under the UDA setting in this field, 
where the action labels are only available on a source dataset, but unavailable on a target dataset for performance evaluation.
This is a more \textit{pragmatic} and \textit{challenging} setting because: 
(1) It is expensive and unfeasible to obtain the annotation of all videos in target dataset from a new environment.
(2) Due to the domain shift, there will be a significant performance drop on the target dataset when directly utilizing the model trained on the source dataset,
which could not be easily handled by simply pre-processing the skeleton-based data (\textit{e.g.,} rotation). See section \ref{section:main_results} for more details.

To this end, we propose a self-supervised learning framework for cross-dataset skeleton-based action recognition under the UDA setting in this paper.
Different from the mainstream UDA methods which apply an adversarial learning based scheme at the \textit{feature level}~\cite{DANN,JAN,CDAN},
our proposed self-supervision scheme concentrates on the \textit{raw data level}, which better preserves their original structure to reduce the domain shift and is easier to implement.
In order to design proper self-supervised learning tasks for skeleton-based action,
we draw lessons from Cubism\footnote{https://en.wikipedia.org/wiki/Cubism}, a famous art genre from the early 20th century, which proposes to deconstruct the object and reassemble the pieces into a screwy yet impressive shape to illustrate the object from different views.
Specially, 
we devise a temporal spatial Cubism strategy, which guides the network to be aware of the permutation of the segments in the temporal domain and the body parts in the spatial domain separately.
During training phase, 
we design the objective function based on two criteria:
(1) minimizing the original action recognition loss on the source domain to improve the discriminative power and 
(2) optimizing the self-supervision loss to enhance the generalization ability.

Moreover, there is a scarcity of available datasets for evaluating UDA approaches for skeleton-based action recognition.
Although some efforts have been made by a recent work~\cite{DBLP:journals/tcsv/gins} on this direction, it still suffers from an obstruction due to the limited data (See Section IV.A for details).
To address this problem, we propose a new experiment setting based on the overlapping action classes of the PKU-MMD~\cite{DBLP:conf/mm/LiuHLS017}, NTU RGB+D~\cite{DBLP:conf/cvpr/ShahroudyLNW16} and Kinetics~\cite{DBLP:journals/corr/KayCSZHVVGBNSZ17}, which are three large-scale and widely used datasets for skeleton-based action analysis.
We conduct experiments on a series of UDA methods and the extensive results on these three datasets as well as other three datasets evaluated in~\cite{DBLP:journals/tcsv/gins}. Extensive experiments have shown that our method sets new state-of-the-art results in this field.

Our main contributions are summarized as follows:
\begin{itemize}
\item[1)] Different from conventional works on skeleton-based action recognition under the fully-supervised paradigm,
we explore a new UDA setting in this realm with greater challenge and more pragmatic value.
\item[2)] Unlike the popular adversarial learning based approaches for UDA, 
we propose a self-supervised learning framework,
which mines the temporal and spatial dependency for skeleton-based sequence and enhance the generalization ability of the model.
\item[3)] In order to facilitate the performance evaluation on this problem, we present a new experiment setting on three large-scale datasets. To our best knowledge, they are currently the largest datasets for cross-dataset skeleton-based action recognition.
\item[4)]
We conduct experiments on six datasets under the setting proposed in this paper and~\cite{DBLP:journals/tcsv/gins}.
Both quantitative and qualitative results demonstrate the superiority of our approach compared with the state of the art. 
\end{itemize}

The remainder of this paper is organized as follows: Section
II briefly reviews some related works. Section III introduces
the proposed approach for cross-dataset skeleton-based action recognition in detail. Section IV reports experimental presents
and analysis, and Section V concludes the paper.

%----------------------------------------------------------
\section{Related Work}
In this section, we briefly review four related topics:
1) skeleton-based action recognition,
2) unsupervised domain adaptation,
3) video-based domain adaptation, and
4) self-supervised learning.

\subsection{Skeleton-based Action Recognition}
Skeleton-based action recognition has attracted growing attention in the realm of computer vision and a variety of methods have been proposed over the past decades.
For a detailed survey we refer the reader to~\cite{DBLP:journals/cviu/HanRHZ17,NTU120,DBLP:journals/cviu/WangLO0E18}, while here we provide a brief literature review.
The early works on skeleton-based action recognition are based on hand-crafted features~\cite{Vemulapalli2014Human,JunwuCVPR17,KoniuszCP16,DBLP:conf/eccv/WangYHLZ16}, 
while recent approaches are devised by designing deep neural networks (DNNs) like convolutional neural networks
(CNNs)~\cite{Liu2017Enhanced,DPRL,DBLP:conf/ijcai/LiZXP18} and recurrent neural networks (RNNs)~\cite{Song2016An,DBLP:conf/iccv/ZhangLXZXZ17,DBLP:conf/eccv/ZhangXLZGZ18}.
In order to better capture the relationship of different joints in the spatial domain or dependency of different frames in the temporal domain, 
a number of works utilized the attention mechanisms~\cite{Song2016An,DBLP:conf/eccv/ZhangXLZGZ18,DBLP:conf/cvpr/SiC0WT19} and 
graph neural networks (GNNs)~\cite{DBLP:conf/aaai/YanXL18,Shi_DGNN_2019_CVPR,Shi_AGCN_2019_CVPR,Li_2019_CVPR,DBLP:conf/cvpr/SiC0WT19} more recently.
Besides, there are various works using both skeleton joints and RGB videos as inputs for action recognition~\cite{PoTion, CMS_Networks, verma2020deep}. For example, Verma \textit{et al.}~\cite{verma2020deep} design two deep neural networks (DNNs) models for the multi-modal inputs respectively, and use a weight product model (WPM) to fuse the softmax scores obtained from the two DNNs.
Different from these works which deal with the input videos from the same dataset during training and testing phases,
we study a more practical and challenging UDA setting to deal with the samples across different datasets.

\subsection{Unsupervised Domain Adaptation}
Reducing the domain shift between the source and target datasets is the core of UDA.
In the past few years, a series of models have been built upon deep neural networks for learning domain-independent representations, 
which show more promising results than early methods based on hand-crafted features~\cite{DBLP:conf/nips/HuangSGBS06,DA_TCA,gong2013connecting}.
Among these, one representative strategy is to design adaptation layers for aligning different distributions~\cite{JAN},
and another popular scheme is to include a domain discriminator sub-network for adversarial learning~\cite{DANN,JAN,MCDUDA,CDAN}.
More recently, there are several attempts on leveraging self-supervised learning for UDA~\cite{DBLP:conf/cvpr/CarlucciDBCT19,DA_SS}.
Under a multi-task learning paradigm,
they optimized the model with the supervision from the source domain, and the auxiliary self-supervision from both source and target domains.
Motivated by the success of these methods in the image domain,
we move a further step in the field of skeleton-based action recognition.
Note that our exploration is non-trivial since the intrinsic structure of the skeleton-based video is quite different from image, and further generalization and adaptation are required.

\subsection{Video-based Domain Adaptation}
Compared with image-based domain adaptation, video-based domain adaptation is a seldom-explored field.
In the literature, a few works have been proposed for RGB videos,
by foreground-weighted histogram decomposition~\cite{HAR_FHD}, 
or performing adversarial learning on the video features~\cite{DBLP:conf/bmvc/JamalNDV18}.  
More recently, Chen et al.~\cite{video_DA} devised TA$^3$N by introducing a temporal relation module and domain attention mechanism.
For skeleton-based video, Tas et al.~\cite{DBLP:conf/bmvc/TasK18} 
and Lin et al.~\cite{DBLP:conf/mm/LinSY020} study the supervised domain adaptation and transfer learning settings, where the action labels of the target dataset are required at the training or fine-tuning stages respectively.
The most relevant work to ours is GINs~\cite{DBLP:journals/tcsv/gins}, which also studied the problem of cross-dataset skeleton-based action recognition under the UDA setting.
In comparison, we proposed a setting with three datasets with larger-scale, and devised a self-supervised learning framework rather the adversarial-based method used in~\cite{DBLP:journals/tcsv/gins}. Experimental results also show the advantage of our method.

\subsection{Self-supervised Learning}
The paradigm of self-supervised learning is to design auxiliary task(s) with the supervision of the data itself,
for example, predicting spatial context~\cite{DBLP:conf/iccv/DoerschGE15} or image rotation~\cite{DBLP:conf/iclr/GidarisSK18}, solving jigsaw puzzles~\cite{DBLP:conf/eccv/NorooziF16} and many others~\cite{DBLP:conf/eccv/ZhangIE16,DBLP:conf/cvpr/LarssonMS17,DBLP:conf/cvpr/PathakKDDE16,DBLP:conf/iccv/DoerschZ17,DBLP:conf/cvpr/He0WXG20,DBLP:journals/corr/abs-2002-05709}.
There have been a number of self-supervised learning methods for RGB videos,
according to the information of ordering~\cite{DBLP:conf/eccv/MisraZH16,DBLP:conf/cvpr/FernandoBGG17,DBLP:conf/iccv/LeeHS017}, geometry~\cite{DBLP:conf/cvpr/GanGLSG18}, correspondence~\cite{wang2019learning,DBLP:conf/cvpr/DwibediATSZ19,DBLP:conf/cvpr/LaiLX20}, motion and appearance statistics~\cite{DBLP:conf/cvpr/WangJBHLL19} or spatio-temporal cubic puzzles~\cite{STCP}.
Compared with these works, besides temporal ordering,
we further explore the relationship of different human body parts for skeleton-based videos by learning from spatial Cubism,
and leverage the advantage of self-supervised learning to seek a better alignment between the source and target domains.

\begin{figure*}[!t]
\includegraphics[width = \linewidth]{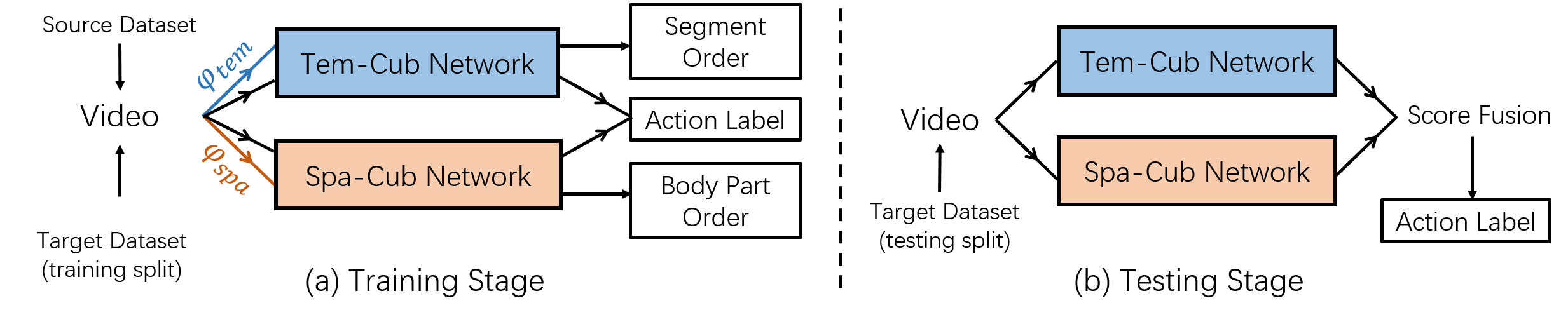} 
\caption{The pipeline of our proposed method. 
(a) During the training stage, the videos from the source and target domains are paired. The videos are randomly chosen for the Cubism transformations and the permuted videos are fed into the networks together with the ordered videos. The network will be optimized according to the Cubism ordering loss from both two domains and the action classification loss only from source domain. The temporal and spatial streams are optimized separately.
(b) During the test stage, the videos from the target domain are fed into the networks to acquire the prediction. The final label is deduced by fusing the predicted scores from two streams.
}
\label{fig:overview}  
\end{figure*}

%----------------------------------------------------------
\section{Approach}
\subsection{Problem Formulation}
We use $\mathcal{D}_s = \{(\boldsymbol{x}_i^s,y_i^s)\}_{i=1}^{n_s}$ to denote a source domain, 
which contains skeleton-based action videos $\{\boldsymbol{x}_i^s\}_{i=1}^{n_s}$ and their action labels $\{y_i^s\}_{i=1}^{n_s}$. Here $i$ denotes the index of the $i$-{th} video, $n_s$ means the number of videos, and the subscript of $n_s$ denotes the source domain.
Similarly, the target domain is defined as $\mathcal{D}_t = \{(\boldsymbol{x}_j^t\}_{j=1}^{n_t}$,
where the action labels are unavailable during the network optimization but can used for the performance evaluation.
Since the videos in the source and target domains are from different datasets,
they correspond to two different joint distributions as $P(\boldsymbol{X}_s,\boldsymbol{Y}_s)$ and $Q(\boldsymbol{X}_t,\boldsymbol{Y}_t)$.
The training should be performed on the source domain with the action labels, and a split of target domain data where the action labels are unavailable.
The testing process is based on the other split of the target domain data which is invisible during the training phase. See section \ref{section:expe} for more details.
There is a more challenging cross-dataset setting which assumes that data from target domain are totally unavailable. The experimental results under this setting are introduced in the section \ref{without_target_domain}.

\subsection{Pipeline Overview}
The motivation of this work is to leverage self-supervised learning and Cubism transformation to reduce the domain shift between the skeleton-based action videos from the source and target datasets.
The concept ``Cubism'' here is originally an art genre from the early 20th century, which breaks and reassembles the objects to convey a greater context.
Inspired by this idea and the progress in self-supervised learning~\cite{DBLP:conf/cvpr/FernandoBGG17,DBLP:conf/iccv/LeeHS017,DBLP:conf/eccv/MisraZH16}, we design two auxiliary pretext tasks, named as temporal Cubism (section 3.3) and spatial Cubism (section 3.4), for skeleton-based action recognition under a new cross-dataset scenario. Accordingly, we devise two networks as the Tem-Cub Network and Spa-Cub Network, where ``Tem-Cub'' and ``Spa-Cub'' are abbreviations of ``temporal Cubism'' and ``spatial Cubism''.
During the training phase, each network is optimized based on one of the self-supervised tasks and the main prediction task jointly.
At the inference period, the final result is obtained by fusing the prediction scores of two networks.
We elaborate on each stage of our pipeline in detail as follows.

\begin{figure}[!t]
\centering
\includegraphics[width =0.85\linewidth]{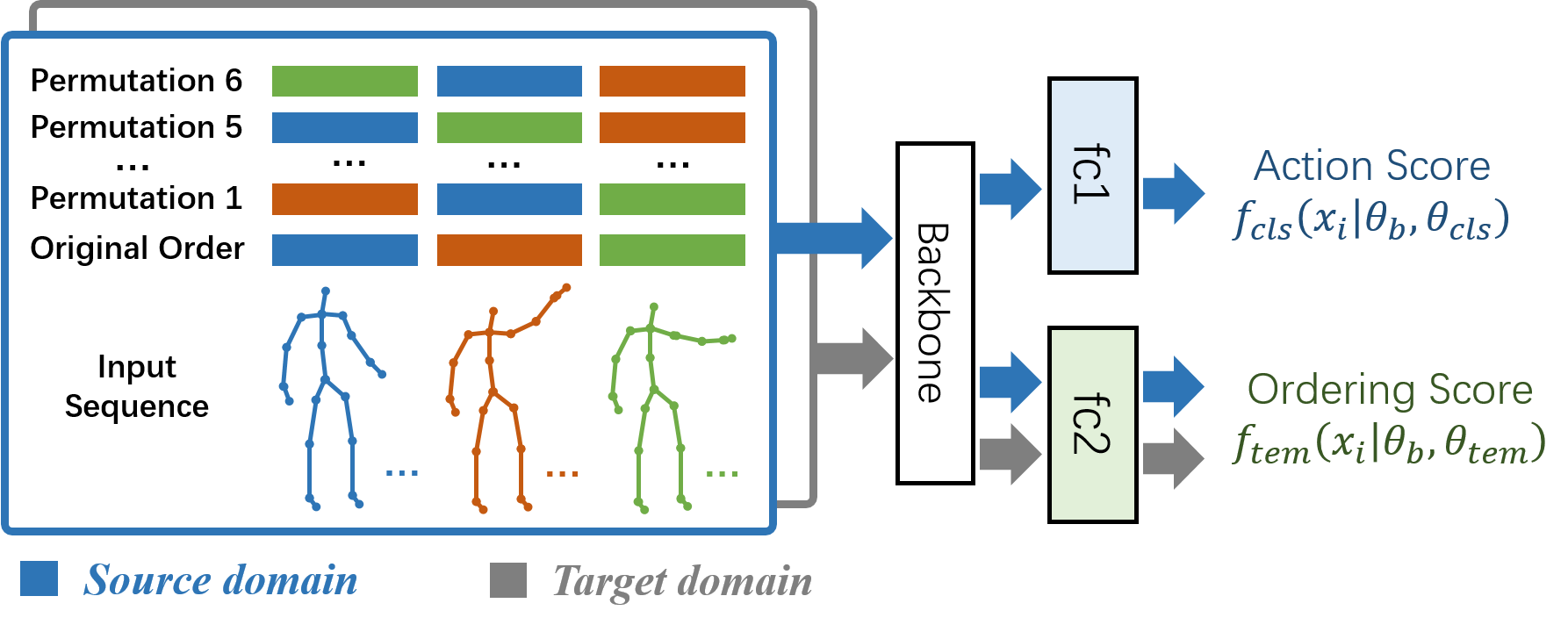} 
\caption{Learning from temporal Cubism. 
Given a video from the source or target domain, 
we divide it into $N$ segments ($N$=3) and permute them to generate a new sample with a new ordering label.
We sent the original data and permuted data into a backbone simultaneously.
The network parameters are optimized in a multi-task learning framework with a total loss of two terms: 
(1) the cross-entropy loss between the predicted action scores and the action labels in the source domain, and
(2) the cross-entropy loss between the predicted ordering scores and the ordering labels in both source and target domains. 
}
\label{fig:tem_jig}  
\end{figure}

\subsection{Learning from Temporal Cubism}
\label{subsection:tem_cub}

Fig. \ref{fig:tem_jig} illustrates our proposed strategy of learning from temporal Cubism. The term temporal Cubism here means we shufﬂe a video in temporal dimension and reorganize them in a new frame order.
Mathematically, we organize each skeleton-based action video as a representation with the size of $F\times K \times D$, where $F$ denotes the number of the frames, $K$ is the number of the joints, and D represents the dimension of joint coordinates.
Given a video sample $\boldsymbol{x}$, we first divide it into $N$ segments uniformly in the temporal domain as
$
\boldsymbol{x} = \begin{bmatrix}(\boldsymbol{x}^{(1)})^T &
	(\boldsymbol{x}^{(2)})^T &
	\dots &
	(\boldsymbol{x}^{(N)})^T\end{bmatrix}^T
$,
where we choose $N=3$ in this paper empirically.
Then a new video $\boldsymbol{x}_{tem}$ with corresponding permutation label $l_{tem}$ in temporal domain is acquired by permuting the segments. This transformation $\varphi_{tem}$ could be presented by a partitioned permutation matrix $\boldsymbol{P}_{tem}$ as:
{\setlength\abovedisplayskip{1pt}
\setlength\belowdisplayskip{1pt}
\begin{equation}
        \boldsymbol{x}_{tem}  = 
        \varphi_{tem}(\boldsymbol{x})  = 
        \boldsymbol{P}_{tem}\boldsymbol{x} .
\end{equation}}

There is only one identity matrix on each row and on each column, and the remaining elements are zero matrices.
For example, if the permutation is to exchange the order of the first and the third segments, the transformation can be written as below:

\begin{equation}
	\boldsymbol{x}_{tem} 
					 = \left[\begin{array}{c c c}
						  \boldsymbol{0} &  \boldsymbol{0} & \boldsymbol{I} \\
						  \boldsymbol{0} &  \boldsymbol{I} &  \boldsymbol{0} \\
						  \boldsymbol{I} &  \boldsymbol{0} &  \boldsymbol{0}
					 \end{array}\right] \begin{bmatrix}
						  \boldsymbol{x}^{(1)} \\
						  \boldsymbol{x}^{(2)} \\
						  \boldsymbol{x}^{(3)}
					  \end{bmatrix} .
\end{equation}

In this way, we build the permuted videos and form the augmented source and target datasets which are presented as follows:
\begin{align}\label{eq:domain}
\mathcal{D}_s^\prime = \{(\boldsymbol{x}_{tem,i}^s, y_i^s, l_{tem,i}^s)\}_{i = 1}^{n_s^\prime}, 
\quad \textit{where} \; \boldsymbol{x}_{tem,i}^s = \boldsymbol{x}^s~\mbox{or}~\boldsymbol{x}_{tem,i}^s = \varphi_{tem}(\boldsymbol{x}^s), \; \boldsymbol{x}^s \in \mathcal{D}_s. \\ \nonumber
\mathcal{D}_t^\prime = \{(\boldsymbol{x}_{tem,i}^t, l_{tem,i}^t)\}_{i = 1}^{n_t^\prime}, 
\quad \textit{where} \; \boldsymbol{x}_{tem,i}^t = \boldsymbol{x}^t~\mbox{or}~\boldsymbol{x}_{tem,i}^t = \varphi_{tem}(\boldsymbol{x}^t), \; \boldsymbol{x}^t \in \mathcal{D}_t.
\end{align}

Here $n_s^\prime$ and $n_t^\prime$ denote the overall number of videos in the augmented source datasets and the augmented target datasets. For the $i$-th video, $y_i$ and $l_{tem,i}$ represent the action label and permutation label of temporal Cubism respectively. Based on the augmented datasets, we design an auxiliary classification task in the temporal domain, which guides the network to learn to recognize the temporal permutation.
During training, the ordered and permuted video samples are packed into batches with a certain ratio which is dependent on a hyper-parameter $p_t$ indicating the percentage of the ordered samples in a batch. This hyper-parameter will be studied in section \ref{subsection:expe_res_and_anal}. Moreover, the permutation way is chosen with equal probability so that transformed videos with different ordering labels are of an identical proportion. 
The mixed batches are fed into a CNN-based backbone (we detail it in section \ref{subsection:implement_detail}) followed by two parallel full-connected classifiers. $f_{cls}$ takes the features of the ordered and disordered samples from the source domain and predicts the action classes while $f_{tem}$ targets to recognize the ordering label for the samples from both source and target domains. Two kinds of losses are computed and combined after the classifiers to optimize the network. 
Comprising two parts of losses, the total loss $\textbf{\textit{J}}_{tem\_total}$ could be formalized as:
\begin{align}\label{eq:tem_loss}
    J_{tem\_total} &= J_c + \lambda_t J_{tem} \nonumber \\
	  &= \frac{1}{n_s^\prime}  \sum_{(\boldsymbol{x}_{tem,i}, y_i) \in \mathcal{D}_s^\prime } J_c(  f_{cls}(\boldsymbol{x}_{tem,i}|\theta_b,\theta_{cls}),y_i) \\
	  &+ \frac{\lambda_t}{n_s^\prime+n_t^\prime} \sum_{(\boldsymbol{x}_{tem,i}, l_{tem,i}) \atop \in \mathcal{D}_s^\prime \cup \mathcal{D}_t^\prime }J_{tem}(  f_{tem}(\boldsymbol{x}_{tem,i}|\theta_b,\theta_{tem}),l_{tem,i}) . \nonumber
\end{align}

Here we adopt the cross-entropy loss for $J_c$ and $J_{tem}$.
$f_{cls}$ and $f_{tem}$ are the softmax scores of action classes and temporal permutation classes.
$\theta_b, \theta_{cls}$ and $\theta_{tem}$ denote the trainable parameters of the backbone, action recognition fc layer and temporal permutation recognition fc layer respectively.
% $\lambda$ is a hyper-parameter we adopt to balance the effective to 
Here $\lambda_t$ is the hyper-parameters to balance the effects of the losses of the main task and self-supervision learning, which will be studied in section \ref{subsection:expe_res_and_anal} as well.

\begin{figure*}[!t]
\includegraphics[width = \linewidth]{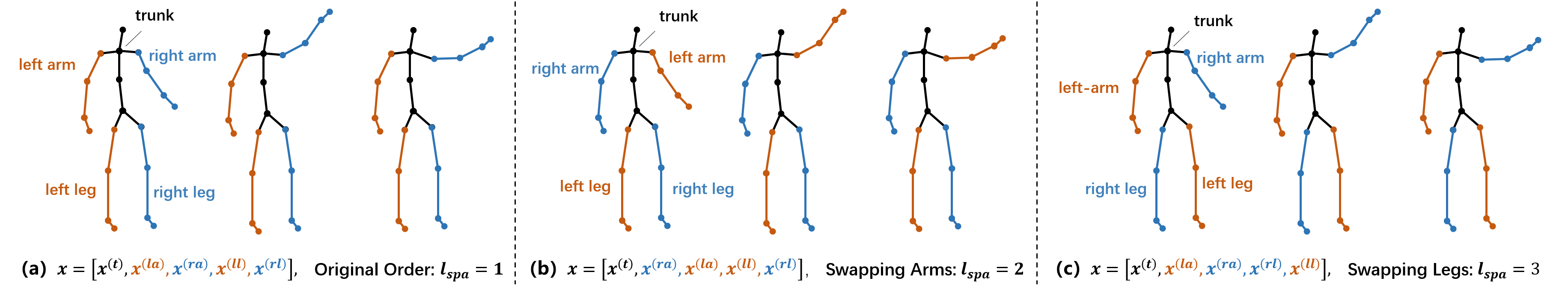} 
\caption{Spatial Cubism.
Given a suite of skeleton, we colored the left part with orange and the right with blue. 
We build the new samples by directly swapping the coordinates of two arms or two legs which results in an uncoordinated pose of the body.
This transformation is implemented by swapping the order of the corresponding elements stored in the linear list.
}
\label{fig:spatial}  
\end{figure*}

\subsection{Learning from Spatial Cubism}
\label{subsection:spa_cub}
As shown in Fig. \ref{fig:spatial},
we design a new self-supervised classification task based on the spatial Cubism among the different body parts.
Specifically, for a skeleton-based action video $\boldsymbol{x}$ defined in the last subsection,
we organize the body parts according to the following ordered list:
$
\boldsymbol{x} = 
\begin{bmatrix}\boldsymbol{x}^{(t)},
    \boldsymbol{x}^{(la)},
    \boldsymbol{x}^{(ra)},
    \boldsymbol{x}^{(ll)},
    \boldsymbol{x}^{(rl)}\end{bmatrix}
$.
The five blocks are corresponding to the data of trunk, left arm, right arm, left leg and right leg respectively.
Similar to the temporal Cubism, 
we can obtain a new sample by performing spatial transformation $\varphi_{spa}$ with another permutation matrix 
$\boldsymbol{P}_{spa}$ as:
\begin{equation}
	\boldsymbol{x}_{spa} = \varphi_{spa}(\boldsymbol{x}) =  \boldsymbol{x}\boldsymbol{P}_{spa} .
\end{equation}

Here we design two concrete instances for $\boldsymbol{P}_{spa}$ as $\boldsymbol{P}_{a}$ and $\boldsymbol{P}_{\ell}$ to swap the coordinates of the joints of the arms and legs respectively:
\begin{equation}
\begin{aligned}
    \boldsymbol{P}_{a} = \left[\begin{array}{ccccc}
                            \boldsymbol{I} &    \boldsymbol{0} &    \boldsymbol{0} &    \boldsymbol{0} &    \boldsymbol{0} \\
                            \boldsymbol{0} &    \boldsymbol{0} &    \boldsymbol{I} &    \boldsymbol{0} &    \boldsymbol{0} \\
                            \boldsymbol{0} &    \boldsymbol{I} &    \boldsymbol{0} &    \boldsymbol{0} &    \boldsymbol{0} \\
                            \boldsymbol{0} &    \boldsymbol{0} &    \boldsymbol{0} &    \boldsymbol{I} &    \boldsymbol{0} \\
                            \boldsymbol{0} &    \boldsymbol{0} &    \boldsymbol{0} &    \boldsymbol{0} &    \boldsymbol{I}
                             \end{array}\right],
                             \quad
    \boldsymbol{P}_{\ell} = \left[\begin{array}{ccccc}
                            \boldsymbol{I} &    \boldsymbol{0} &    \boldsymbol{0} &    \boldsymbol{0} &    \boldsymbol{0} \\
                            \boldsymbol{0} &    \boldsymbol{I} &    \boldsymbol{0} &    \boldsymbol{0} &    \boldsymbol{0} \\
                            \boldsymbol{0} &    \boldsymbol{0} &    \boldsymbol{I} &    \boldsymbol{0} &    \boldsymbol{0} \\
                            \boldsymbol{0} &    \boldsymbol{0} &    \boldsymbol{0} &    \boldsymbol{0} &    \boldsymbol{I} \\
                            \boldsymbol{0} &    \boldsymbol{0} &    \boldsymbol{0} &    \boldsymbol{I} &    \boldsymbol{0}
                             \end{array}\right] .
\end{aligned}
\end{equation}

Through these transformations, the skeleton-based video would convey the screwy actions which refer to the spatial Cubism. 
By learning to discover these,
the network would have a better generalization ability on the spatial domain.
Similar to that of the temporal Cubism, we construct the augmented source dataset $\mathcal{D}''_s$ and target dataset $\mathcal{D}''_t$ as follows:
\begin{align}\label{eq:domain_spa}
\mathcal{D}''_s = \{(\boldsymbol{x}_{spa,i}^s, y_i^s, l_{spa,i}^s)\}_{i = 1}^{n''_s}, 
\quad \textit{where} \; \boldsymbol{x}_{spa,i}^s = \boldsymbol{x}^s~\mbox{or}~\boldsymbol{x}_{spa,i}^s = \varphi_{spa}(\boldsymbol{x}^s), \; \boldsymbol{x}^s \in \mathcal{D}_s. \\ \nonumber
\mathcal{D}''_t = \{(\boldsymbol{x}_{spa,i}^t, l_{spa,i}^t)\}_{i = 1}^{n''_t}, 
\quad \textit{where} \; \boldsymbol{x}_{spa,i}^t = \boldsymbol{x}^t~\mbox{or}~\boldsymbol{x}_{spa,i}^t = \varphi_{spa}(\boldsymbol{x}^t), \; \boldsymbol{x}^t \in \mathcal{D}_t .
\end{align}

We introduce a hyper-parameter $p_s$ to indicate the percentage of the ordered samples in a batch during the training phase.
The total loss $\textbf{\textit{J}}_{spa\_total}$, in this case, could be formalized as:
\begin{equation} \label{eq:spa_loss}
    \begin{aligned}
        J_{spa\_total} &= J_c + \lambda_s J_{spa} \\
		  &= \frac{1}{n''_s} \sum_{(\boldsymbol{x}_{spa,i}, y_i) \in \mathcal{D}''_s} J_c(f_{cls}(\boldsymbol{x}_{spa,i} | \theta_b, \theta_{cls}), y_i) \\
		  &+ \frac{\lambda_s}{n''_s+n''_t} \sum_{(\boldsymbol{x}_{spa,i}, l_{spa,i}) \atop \in \mathcal{D}''_s \cup \mathcal{D}''_t} J_{spa}(f_{spa}(\boldsymbol{x}_{spa,i} | \theta_b, \theta_{spa}), l_{spa,i}) .
    \end{aligned}
\end{equation}

 The variables in Equation (\ref{eq:domain_spa}) and (\ref{eq:spa_loss}) have the similar definitions with those in Equation (\ref{eq:domain}) and (\ref{eq:tem_loss}). 
We present a mathematical algorithm of our method in Algorithm \ref{alg:train TS-Cub}.

\begin{algorithm}[t]
	\DontPrintSemicolon
	\KwIn{skeleton-based videos from source domain $\mathcal{D}_s = \{(\boldsymbol{x}_i^s,y_i^s)\}_{i=1}^{n_s}$ and target domain $\mathcal{D}_t = \{(\boldsymbol{x}_j^t\}_{j=1}^{n_t}$, training epoch $\Gamma$.\;
	}
	\KwOut{The weights of the Tem-Cub Network $\theta_t$ and Spa-Cub Network $\theta_s$.}
	\SetAlgoLined	
	// \emph{Training the Tem-Cub Network:} \; 
	Perform temporal transformation to obtain $\mathcal{D}'_s$, $\mathcal{D}'_t$ based on $\mathcal{D}_s$, $\mathcal{D}_t$ and Eqn. (\ref{eq:domain}). \;
	Initialize $\theta_t$.\;
	\For{$k \leftarrow$ 1, 2, ..., $\Gamma$}
{
	Feed the input data through Tem-Cub Network. \;
	Calculate the objective function $\textbf{\textit{J}}_{tem\_total}^k$ at the $k$-th epoch by Eqn. (\ref{eq:tem_loss}). \;
    Update $\theta_t$ by back propagation.\;
}    
	// \emph{Training the Spa-Cub Network:} \; 
	Perform spatial transformation to obtain $\mathcal{D}''_s$, $\mathcal{D}''_t$ based on $\mathcal{D}_s$, $\mathcal{D}_t$ and Eqn. (\ref{eq:domain_spa}). \;
	Initialize $\theta_s$.\;
	\For{$k \leftarrow$ 1, 2, ..., $\Gamma$}
{
	Feed the input data through Spa-Cub Network. \;
	Calculate the objective function $\textbf{\textit{J}}_{spa\_total}^k$ at the $k$-th epoch by Eqn. (\ref{eq:spa_loss}). \;
    Update $\theta_s$ by back propagation.\;
}   
\textbf{Return:} The parameters $\theta_t$ and $\theta_s$ of the Tem-Cub Network and Spa-Cub Network.
\caption{Training Procedure of Our TS-Cub} \label{alg:train TS-Cub}	
\end{algorithm}

\subsection{Two-Stream Fusion}
In order to further boost performance, we explore several approaches to couple the temporal and spatial Cubism transforms. One is to apply the two kinds of transforms simultaneously and therefore divide the videos into finer-grained atoms (see section \ref{subsection:expe_res_and_anal} for details). However, this results in a more complex task, which might bring more difficulty for optimizing the network and more cost in data pre-processing.

Though feature-level fusion is a common two-stream fusion strategy, we do not apply it in this paper.
This is because spatial and temporal streams implement different auxiliary tasks, and feature-level fusion will make it much more difficult to recognize the ordering label.
Actually, as explored by several previous works~\cite{DBLP:conf/nips/SimonyanZ14,DBLP:conf/eccv/WangXW0LTG16}, it is more effective and efficient to separately deal with the temporal and spatial information and combine them after the network.
Hence, we explore several approaches to fuse softmax scores from the temporal and spatial streams during the inference stage, \textit{e.g.,} Weighted Arithmetic Mean (WAM), Weighted Root Squared Mean (WRSM), Weighted Geometric Mean (WGM) and Max Pooling (MP). The experimental results and more details are shown in Table \ref{tab:other_self_super_tasks} in later section.

%----------------------------------------------------------
\section{Experiments}

\label{section:expe}
\subsection{Datasets and Experiment Settings} 
To our best knowledge, there are very few benchmarks for cross-dataset skeleton-based action recognition. Although the recent work~\cite{DBLP:journals/tcsv/gins} has proposed two benchmarks for this problem, they mainly have two drawbacks. The first is the scales of these datasets are relatively small, and the second is that it adopts a ``test-time training'' strategy~\cite{SunICML20}, which utilizes the test data (without using their action labels) during training. This might be not practical in some real-world scenarios.

\begin{table*}[!t]
    \small
	\centering
	\caption{Comparison of the proposed unsupervised domain adaptation setting for skeleton-based action recognition with that of the previous work~\cite{DBLP:journals/tcsv/gins}. Here ``Nm'' (m = 8, 51, 12) denotes m-action subset of NTU RGB+D.}
	\setlength{\tabcolsep}{7pt}
	\begin{tabular}{l | c c c c |c c }
	\hline
	 & \multicolumn{4}{c|}{Training} &  \multicolumn{2}{c}{Testing}\\
	 & Source& \# Clips & Target & \# Clips & Target & \# Clips\\ 
	\hline
	NTU$\to$SBU~\cite{DBLP:journals/tcsv/gins} & N8 & 7513 & SBU & 282 & SBU & 282 \\
	ORGBD$\to$MSRA3D~\cite{DBLP:journals/tcsv/gins} & ORGBD & 240 &MSRA3D & 100 & MSRA3D & 100 \\
	 \hline
	P$\to$N-CV (Ours) & P & 21544  & N51-CV-train & 31989 & N51-CV-test& 16092\\
	P$\to$N-CS (Ours) & P &21544 & N51-CS-train & 34068 & N51-CS-test& 14013\\
	N$\to$P-CV (Ours) &  N51 & 48081 & P-CV-train & 14356 & P-CV-test&7188 \\
	N$\to$P-CS (Ours) &  N51 & 48081 &  P-CS-train & 18840 & P-CS-test&2704 \\
	\hline
	N$\to$K (Ours) & N12 & 11256 & K-train& 8912 & K-test & 787 \\
	K$\to$N (Ours) & K & 9699 &N12-CV-train&7476 & N12-CV-test & 3780 \\
	\hline
	\end{tabular}
	\label{tab:eval_pro}  
\end{table*}

To address these issues, 
we define two groups of experiment settings crossing three large-scale datasets in this paper: NTU RGB+D~\cite{DBLP:conf/cvpr/ShahroudyLNW16}, PKU-MMD~\cite{DBLP:conf/mm/LiuHLS017}, and Kinetics~\cite{DBLP:journals/corr/KayCSZHVVGBNSZ17}.
We present a comparison of the proposed settings with the previous work~\cite{DBLP:journals/tcsv/gins} in Table \ref{tab:eval_pro}. We detail the corresponding datasets and the experimental settings as follows.

\noindent \textbf{NTU RGB+D}: The NTU RGB+D dataset is a large-scale
dataset for evaluating skeleton-based action recognition.
The dataset contains 56,880 skeleton-based videos of
60 action categories. There are two evaluation protocols as
cross-subject (CS) and cross-view (CV). Under CS setting,
there are videos of 40 subjects used for training while the
videos of the rest 20 subjects are used for the test. Under
CV setting, The videos from view 2 and 3 are used for
training, while the videos from view 1 are used for the test.

\noindent \textbf{PKU-MMD}: There are 1,076 long videos in the PKU-MMD
dataset, which is originally presented for skeleton-based
action detection. In order to construct a dataset for
unsupervised domain adaptation on action recognition, we
trim each long video according to their temporal annotations
and obtain 21,544 video clips of 51 actions. Similar to
the NTU RGB+D dataset, the cross-subject and cross-view
settings are recommended for PKU-MMD dataset. Under
CS setting, there are 18,840 training videos and 2,704 test
videos. Under CV setting, there are 14,356 training videos
and 7,188 test videos.

\noindent \textbf{Kinetics}: Kinetics is a large-scale dataset for action
recognition containing about 300,000 video clips collected from Youtube. Each clip in Kinetics contains around 300
frames. These video clips cover 400 action categories and
under each category, there are more than 400 samples for
training and about 100 for the test. The original Kinetics
dataset releases only raw RGB sequences. Hence we
adopt the estimated poses provided by~\cite{DBLP:conf/aaai/YanXL18} extracted by
OpenPose~\cite{CaoZhe2018_OpenPose} to study the skeleton-based UDA task.

\noindent \textbf{P$\leftrightarrow$N.} We perform unsupervised domain adaptation between PKU-MMD and NTU RGB+D. 51 action categories are extracted from NTU RGB+D to pair with the actions in PKU-MMD.
Both CV and CS settings are adopted for evaluation.
For clarification, we use \textit{N51} to denote the 51-action subset of NTU RGB+D and \textit{P} to denote PKU-MMD. The infixes \textit{CV}, \textit{CS} and suffixes \textit{train}, \textit{test} are used to indicate the subset, \textit{e.g.}, \textit{N51-CS-train} implies the training set of NTU RGB+D under cross-subject setting. Due to the limited space, we show the paired action classes of P$\leftrightarrow$N in our project page.

\noindent \textbf{N$\leftrightarrow$K.} Experiments are carried out between NTU RGB+D and Kinetics as well. 
We select 12 paired actions from NTU RGB+D and Kinetics for domain adaptation. As the estimated pose data on Kinetics are 2-dimensional, we extract the coordinates of x and y axes from NTU RGB+D to get a similar 2D skeleton.
The Kinetics subset is partitioned into the training and test subsets in accordance with the raw division while NTU RGB+D is used under only CV setting. 
Similarly, the subset of NTU RGB+D is marked as \textit{N12} and Kinetics is marked as \textit{K}. The suffixes \textit{train} and \textit{test} are used to indicate the subset as well.
Same as before, the paired action classes of N~\cite{DBLP:conf/cvpr/ShahroudyLNW16}$\leftrightarrow$K~\cite{DBLP:journals/corr/KayCSZHVVGBNSZ17} are presented in the project page.

In order to make a better evaluation of our method, we also conduct experiments on the 
SBU Kinect Interaction dataset (SBU)~\cite{kiwon_hau3d12}, 
Online RGBD Action dataset (ORGBD)~\cite{DBLP:conf/accv/YuLY14} and MSRDaily Activity3D dataset (MSRDA3D)~\cite{DBLP:conf/cvpr/WangLWY12}, following the same settings proposed in the previous work~\cite{DBLP:journals/tcsv/gins}.
The experimental results and analysis are described in detail as below.

\subsection{Compared Methods} In the following section, we first conduct experiments and acquire the results on \textit{Source Only} and \textit{Target Only}.
\textit{Source Only} indicates a baseline method which trains a model in the source domain, and directly evaluates the testing data on the target domain without supervision.
\textit{Target Only} denotes to utilize the ground-truth action labels in the target domain for training, which provides an upper bound for this problem. 
Besides, because there are very few models designed for skeleton-based action recognition under the UDA setting, we compare our model with some image-based UDA models (\textit{i.e.}, MMD, DANN, JAN, CDAN, BSP, MCC) and a RGB video-based model (\textit{i.e.}, TA$^{3}$N).
We replace those models' backbone with HCN and apply the same experimental setting with our model for fair comparison.
Specifically, for TA$^{3}$N, besides replacing the feature extractor with HCN, we add a fully-connected layer between HCN and the spatial-temporal adversarial module in TA$^{3}$N to make them compatible.
Moreover, we also compare our method with GINs, which is the most related work on cross-dataset skeleton-based action recognition.

In our paper, there are mainly three kinds of information to be utilized for cross-dataset transfer. They are temporal information (T), spatial information (S) and their combination (TS), which are used for our Tem-Cub Network, Spa-Cub Network and TS-Cub Network respectively.
For the compared methods, most of them perform transfer based on the temporal-spatial feature extracted by HCN backbone (TS), except the GINs~\cite{DBLP:journals/tcsv/gins} focused on the relation of different joints at the spatial domain (S).

\begin{table}[t]
	\centering 
	\caption{
	Study on the PReLU activation function and Rotation processing.
	Source domain: PKU-MMD (P) dataset. Target domain: NTU RGB+D (N51) .}
	\begin{tabular}{l c c}
	\hline
	Method & P$\to$N-CV & P$\to$N-CS\\
	\hline
	 Source Only~\cite{DBLP:conf/ijcai/LiZXP18}   & 51.9 & 47.9\\  
	HCN~\cite{DBLP:conf/ijcai/LiZXP18}  & 50.9 & 45.8\\
	\hline
	HCN + PReLU & 51.9 & 47.9\\
	HCN + Rot. & 53.4 & 50.3 \\
	HCN + Rot. + PReLU & \textbf{54.9} & \textbf{50.5} \\
	\hline
	\end{tabular}
	\label{table:baseline}  
\end{table}

\subsection{Implementation Details}
\label{subsection:implement_detail}
We conduct experiments on a system with the Intel(R) Xeon(R) CPU E5-2698 v4 @ 2.00Ghz.
We implement our method with the PyTorch toolbox and train the model on Nvidia GTX 1080 Ti GPU. The training epoch $\Gamma$ is set to 400 for both Tem-Cub and Spa-Cub.
We adopt the HCN~\cite{DBLP:conf/ijcai/LiZXP18} as the backbone because of its effectiveness and efficiency.
In order to ameliorate the performance, we made two modifications.
The PReLU~\cite{DBLP:conf/iccv/HeZRS15} is used as the non-linear activation function instead of the commonly used ReLU and
a rotation pre-processing is applied before the networks to eliminate the difficulty for bridging the domain gap induced by the videos captured from different views.
specifically, we rotate the skeleton around the z axis to make the connection of the joints \textit{right shoulder} and \textit{left shoulder} parallel with the x axis and the connection of \textit{spine base} and \textit{spine} parallel with y axis.

\subsection{Experimental results}
\label{section:main_results}

\begin{table}[t]
	\centering
	\caption{Comparison of the skeleton-based action recognition accuracy (\%) under the UDA setting between PKU-MMD (P) and NTU RGB+D (N51) datasets.}
	\setlength{\tabcolsep}{2.3pt}
	\begin{tabular}{l|cccc|c}
		\hline
		Method &	P$\to$N-CV & P$\to$N-CS &	N$\to$P-CV& N$\to$P-CS & year\\
		\hline
		Target Only & 91.3 & 84.6 &	95.7 & 93.1 & -\\
		Source Only & 54.9 & 50.5 &	59.6 & 57.6 & -\\
		\hline
		MMD~\cite{DBLP:journals/corr/Long015} &	55.4 & 51.7 &	61.3 & 59.4 & 2015\\
		DANN~\cite{DANN} & 58.1 & 52.4 & 61.9 & 58.8 & 2017\\
		JAN~\cite{JAN} & 51.9 & 47.1 & 65.3 & 63.1 & 2017\\
        CDAN~\cite{CDAN} & 54.9 & 51.1 & 63.8 & 61.3 & 2018\\
        BSP~\cite{BSP} & 55.7 & 49.3 & 64.0 & 63.0 & 2019\\
        TA$^3$N~\cite{video_DA} & 55.9 & 51.2 & 66.2 & 65.9 & 2019\\
        GINs~\cite{DBLP:journals/tcsv/gins} & 44.9 & 40.0 & 54.7 & 52.4 & 2020\\
        MCC~\cite{DBLP:conf/eccv/JinWLW20}   & 56.1 & 52.2 & 64.2 & 61.7 & 2020 \\
		\hline
		Tem-Cub (Ours) &	58.3&52.5 &	65.2 & 62.8 & \\
		Spa-Cub (Ours) &	56.5&51.0 & 61.2&59.3  & \\
		TS-Cub (Ours) &		\textbf{59.2}&\textbf{53.6} &	\textbf{65.5}&\textbf{63.3}  & \\
		\hline
	\end{tabular}
	\label{tab:SAR_SOTA}
\end{table}

\noindent \textbf{Analysis of the Baseline Model:}
We adopt the HCN\cite{DBLP:conf/ijcai/LiZXP18} as the backbone and make two modifications that are introduced in the last subsection. 
As shown in Table \ref{table:baseline}, both these two modifications manage to ameliorate the performance of HCN. Combining PReLU and rotation pre-processing could further improve the performance. Hence we will refer baseline to the improved HCN in the sequel.
Though pre-processing can reduce the domain gap to some extent, there are still some phenomena that could not be easily handled. For example, bridging the gap between 2D and 3D skeleton-based videos.
Moreover, conventional pre-processing methods, \textit{e.g.,} rotation, scale normalization, \textit{etc}, are executed based on several specific joints. 
But in skeleton-based videos, these joints might be inaccurate or missing (padding 0).
Actually, these are common phenomena in skeleton-based datasets.
In this case, performing pre-processing on these data might even cause negative effects.
Therefore, our proposed method owns greater generalization ability than pre-processing.

\begin{figure*}[t]
\includegraphics[width =\linewidth]{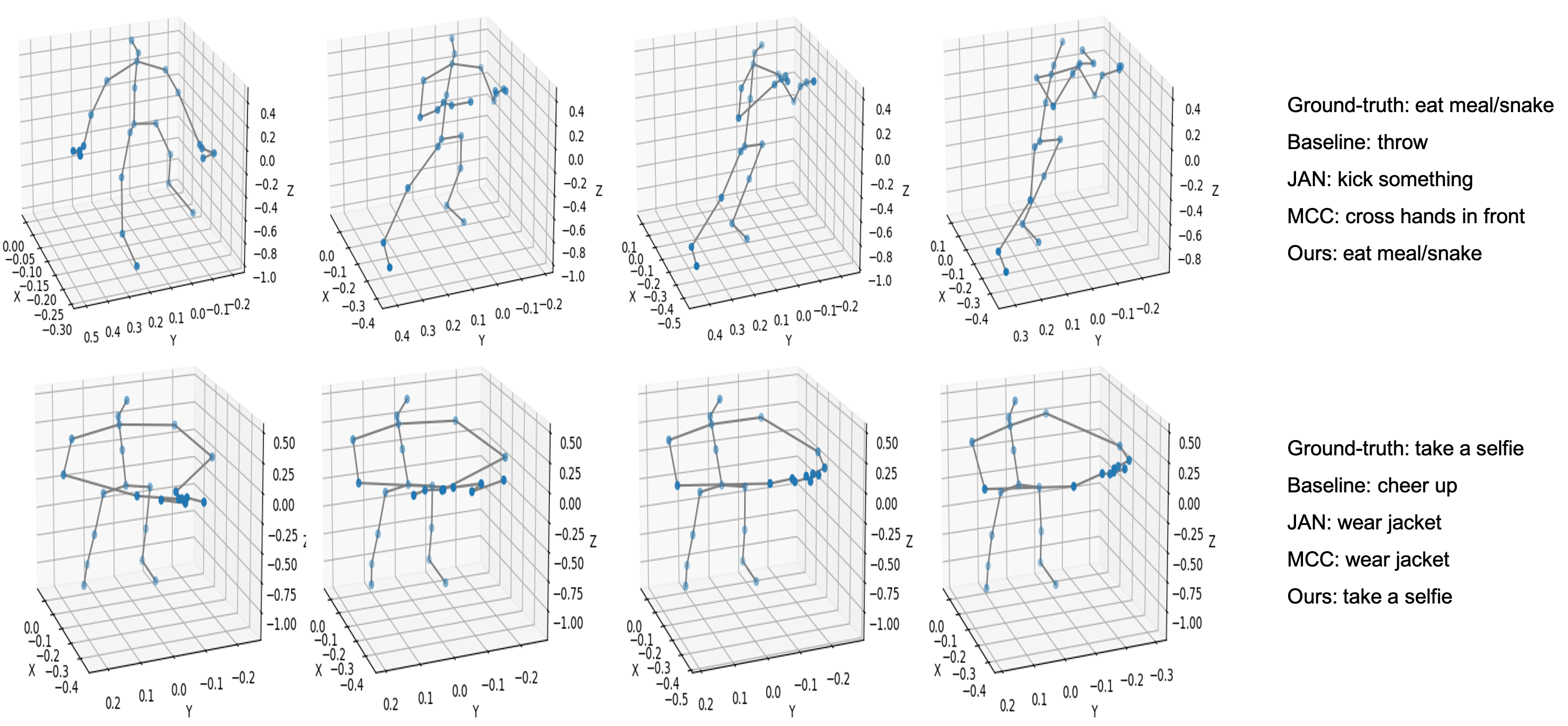}
\caption{Visualization of two skeleton-based videos under the N$\to$P-CV setting and the predicted results of different methods.}
\label{fig:case_visual}
\end{figure*}

\noindent \textbf{Evaluation on PKU $\leftrightarrow$ NTU:}
We show our results compared to the state-of-the-arts under the P$\to$N and N$\to$P settings in Table \ref{tab:SAR_SOTA}.
We first observe the large difference between \textit{Target Only} and \textit{Source only} caused by the domain shift (around 35\% accuracy drops), 
which indicates the greater challenge of the UDA setting compared with conventional fully supervised-learning setting.
Moreover, according to the results, our proposed method acquires consistent improvement of the performance of skeleton-based unsupervised domain adaptation.
To be specific, it is shown that our TS-Cub method exceeds the baseline by 4.3\% and 3.1\% under CV and CS settings respectively and achieves remarkable improvement based on the state-of-the-arts. The performance is 3.8\%, 1.1\%, 7.3\%, 4.3\%, 3.5\%, 3.1\% higher than that of MMD, DANN, JAN, CDAN, BSP and MCC in order under P$\to$N-CV setting.
Meanwhile, our method constantly outperforms the  state-of-the-art video domain adaptation model TA$^{3}$N by 4.4\%, 2.7\%, 1.5\%, 0.7\% under four P$\leftrightarrow$N settings.
Figure~\ref{fig:case_visual} shows two failure cases on compared methods but succeeded on TS-Cub. For the two skeleton-based videos under the N$\to$P-CV setting, the ground-truth labels are \textit{eat meal/snake} and \textit{take a selfie}, while the compared methods give wrong predictions, our TS-Cub approach obtains the right action labels, which demonstrate its effectiveness for cross-dataset action recognition.
Besides, it is observed that our Tem-Cub method performs consistently better than Spa-Cub. We explain this fact by that it is a little difficult to discriminate between the ordered and permuted instances keeping strong spatial symmetry, which affects the performance of the Spa-Cub method.
As conclusion, our proposed Cubism methods have the capability to improve the domain adaptation performance. Tem-Cub appears to be more effective than Spa-Cub and combining the information from both two streams will further enhance recognition accuracy.

\begin{table}[t]
	\centering 
	\caption{Comparison of the skeleton-based action recognition accuracy (\%) under the UDA setting.
	Source domain: NTU RGB+D (N12) dataset. Target domain: Kinetics (K) dataset.}
	\begin{tabular}{l c c c}
	\hline
	Method & N$\to$K & K$\to$N & Year\\
	\hline
	Target Only & 40.8 & 89.1 & -\\
	Source Only & 14.4 & 22.5 & -\\  
	\hline
	MMD~\cite{DBLP:journals/corr/Long015} & 14.9  & 22.8 & 2015\\
	DANN~\cite{DANN}& 16.5   & 22.9 & 2017\\
	JAN~\cite{JAN} & 15.0    & 24.4 & 2017\\
	CDAN~\cite{CDAN} & 14.9    & 24.0 & 2018\\
	BSP\cite{BSP} & 14.6 & 16.6  & 2019\\
	TA$^{3}$N~\cite{video_DA} & 15.6 & 25.6  & 2019\\
	GINs~\cite{DBLP:journals/tcsv/gins} & 16.4 & 23.4 & 2020 \\
	MCC~\cite{DBLP:conf/eccv/JinWLW20}   & 15.2 & 26.2 & 2020 \\
	\hline
	Tem-Cub (Ours) & 16.4 & 29.3 & 2021\\
	Spa-Cub (Ours) & 14.6 & 22.9 & 2021\\
	TS-Cub (Ours) & \textbf{16.8} & \textbf{29.6} & 2021 \\
	\hline
	\end{tabular}
	\label{table:SAR_SOTA_N&K}  
\end{table}
\begin{figure*}[!t]
\includegraphics[width = \linewidth]{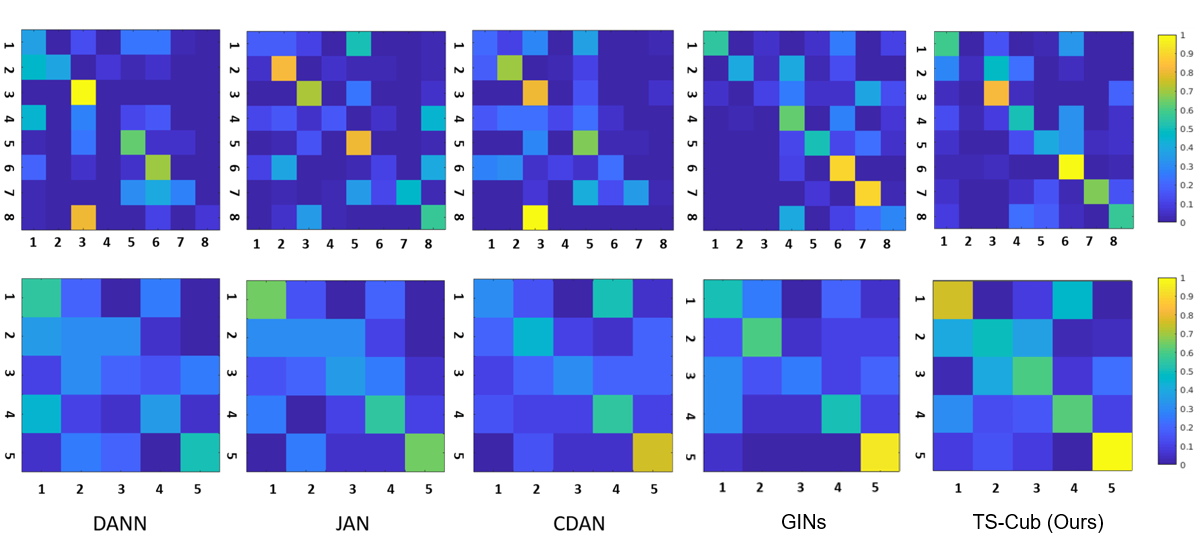}
\caption{Visualization of confusion matrices.
We show the ground truth labels and the predicted labels on the vertical axis and the horizontal axis respectively.
The first row displays the results of NTU $\to$ SBU, where the labels are: punching (1), exchanging something (2), hugging (3), handshaking (4), pushing (5), kicking (6), walking apart (7), walking towards (8).
The second row represents the results of ORGBD $\to$ MSRDA3D, where the labels are: drinking (1), eating (2), using laptop (3), making phone call (4), reading book (5).
}
\label{fig:conf}  
\vspace{0.1cm}
\end{figure*}

\begin{table}[t]
\centering 
\caption{Comparison of the skeleton-based action recognition accuracy (\%) under the unsupervised domain adaptation setting. * Since the target data is not divided into subsets for training and testing~\cite{DBLP:journals/tcsv/gins}, we could not evaluate the ``Target Only'' in this table.}
\vspace{-0.35 cm}
\begin{tabular}{l c c c}
\hline
Method & NTU $\to$ SBU & ORGBD $\to$ MSRDA3D & Year\\
\hline
Target Only*   & - & - & -\\  
Source Only   & 35.8 & 48.3 & -\\  
\hline
MMD~\cite{DBLP:journals/corr/Long015} &31.4 & 25.5 & 2015 \\
DANN~\cite{DANN} & 46.3  & 39.3 & 2017\\
JAN~\cite{JAN} &47.6 & 49.2 & 2017\\
CDAN~\cite{CDAN} &39.9 & 48.7  & 2018\\
GAKT~\cite{DBLP:conf/eccv/DingLSF18} & 31.8 & 48.4 & 2018 \\
BSP~\cite{BSP} & 32.4 & 41.3 & 2019\\
GINs~\cite{DBLP:journals/tcsv/gins} & 50.7 & 51.5 & 2020 \\
\hline
Tem-Cub (Ours) & 50.7 & 52.5 & 2021 \\
Spa-Cub (Ours) & 46.8 & \textbf{53.1} & 2021  \\
TS-Cub (Ours)  & \textbf{51.1} & 53.0 & 2021 \\
\hline
\end{tabular}
\label{table:SAR_SOTA_S_M}  
\end{table}

\noindent \textbf{Evaluation on NTU $\leftrightarrow$ Kinetics:}
We conducted further experiments on NTU RGB+D and Kinetics datasets to verify the effectiveness of our methods and present the results in Table \ref{tab:SAR_SOTA}. 
Compared to P$\leftrightarrow$N, there is an even larger performance gap between \textit{Target Only} and \textit{Source only} in N$\leftrightarrow$K.
As for our proposed methods,
TS-Cub exceeds the baseline by 2.4\% and 7.1\% under N$\to$K and K$\to$N settings respectively and exceeds the best one from the state-of-the-arts.
Besides, it is noticed that the adaptation performance from Kinetics to NTU RGB+D is significantly higher than that from NTU RGB+D back to Kinetics. This should be attributed to the features of the two underlying datasets.
Kinetics is captured from a wider domain, the Youtube videos in the wild, while NTU RGB+D is collected under the artificially predefined settings.
Accordingly, Kinetics conveys more information that might facilitate the adaptation and NTU RGB+D carries less noise which brings less difficulty for recognition.
For the aforementioned reasons, an adaptation from a more complex domain could result in relatively higher performance. This phenomenon holds for the adaption between PKU-MMD and NTU RGB+D as well.

\label{subsection:expe_res_and_anal}

\noindent \textbf{Evaluation on NTU $\to$ SBU:}
We then conducted experiments between NTU and SBU datasets, following the setting in ~\cite{DBLP:journals/tcsv/gins}. 
We present the results in Table \ref{table:SAR_SOTA_S_M}, where the performances of other methods are reported in \cite{DBLP:journals/tcsv/gins}.
As a result, we can find that most of the methods can boost the accuracy compared with  \textit{Source Only}, and our TS-Cub method achieves the highest accuracy of 51.1\%.
We further show the confusion matrices of different methods in the top row of Fig. \ref{fig:conf}. Our ST-Cubism shows strong performance on the actions of \textit{kicking} and \textit{hugging}.

\noindent \textbf{Evaluation on ORGBD $\to$ MSRDA3D:}
Table \ref{table:SAR_SOTA_S_M} shows the experimental results on the ORGBD $\to$ MSRDA3D setting~\cite{DBLP:journals/tcsv/gins}.
Compared with other aforementioned datasets, ORGBD $\to$ MSRDA3D dataset is rather small and only contains 5 categories. 
Referring to Table \ref{table:SAR_SOTA_S_M}, we can find that \textit{Source Only} exceeds almost all the compared methods. This may attribute to the fact that adversarial learning methods require numerous training data. Meanwhile, the proposed methods achieve the results of 52.5\% (Tem-Cub), 53.1\% (Spa-Cub) and 53.0\% (TS-Cub) respectively, surpassing all the other compared methods. This shows the robustness of our methods to the fewer training data in comparison with the mainstream adversarial methods.

We display the compared confusion matrices in the bottom row of Fig. \ref{fig:conf}. Our method could recognize the action of \textit{reading book} well, but would be sometimes confused by the action of \textit{eating} and \textit{using laptop}.

\begin{table}[t]
	\centering
	\caption{
	Comparison of the averaged computational cost (ms) based on a single video during the inference period. 
	Experiments are conducted under the P$\to$N-CV setting.} 
	\vspace{-0.35 cm}
	\setlength{\tabcolsep}{0.7pt}
	\begin{tabular}{l|cccccc||cc|c}
		\hline
		Method & MMD~\cite{DBLP:journals/corr/Long015} & DANN~\cite{DANN} & JAN~\cite{JAN} & CDAN~\cite{CDAN} & BSP~\cite{BSP} & GINs~\cite{DBLP:journals/tcsv/gins} & Tem-Cub & Spa-Cub & TS-Cub \\	
        \hline
        Time & 0.106 & 0.153 & 0.083 & 0.093 & 0.305 & 0.241 & 0.112 & 0.109 & 0.211 \\
		\hline
	\end{tabular}
	\label{tab:SAR_cost}
\vspace{-0.1 cm}	
\end{table}

\noindent \textbf{Evaluation of the Computational Cost:}
We run experiments under the P$\to$N-CV setting and report the averaged computational cost based on a single video in Table \ref{tab:SAR_cost}. As it shows, our Tem-Cub network and Spa-Cub network cost 0.112 ms and 0.109 ms to predict the action label of each single video, achieving comparable speed with the fastest method JAN~\cite{JAN}. Though our final model TS-Cub requires more time than some of the compared methods, it can still satisfy the real-time application requirement.

\begin{figure*}[t]
\includegraphics[width = 0.8\linewidth]{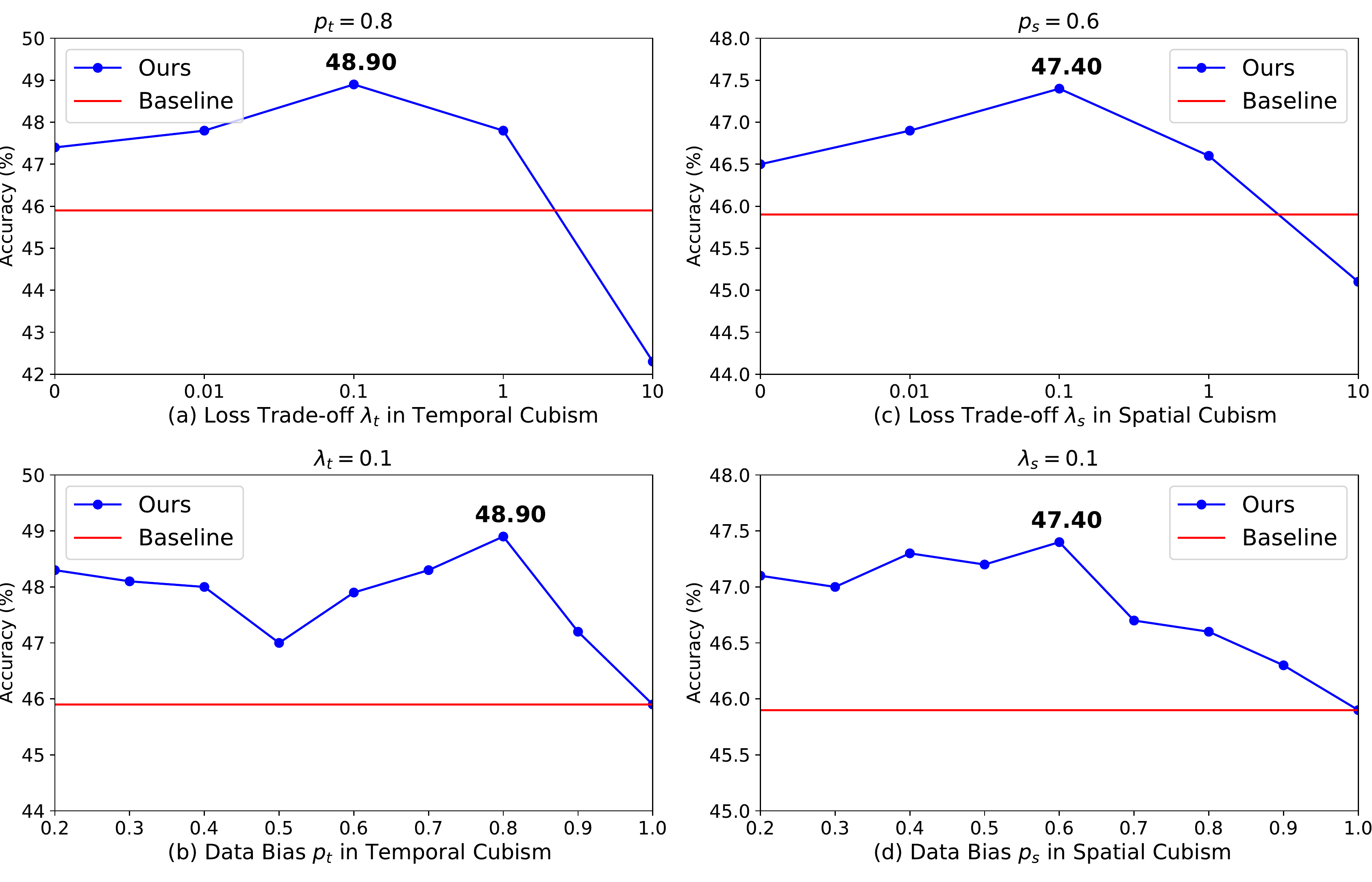}
\caption{The ablation study of the hyper-parameters, the loss trade-off parameter $\lambda$ and data bias parameter $p$, in our Cubism methods.
}
\label{fig:hyper}
\vspace{-0.1cm}
\end{figure*}

\subsection{Analysis of the TS-Cub}
\noindent \textbf{Analysis of the Hyper-parameters:}
We studied the impact of the hyper-parameters in our method, the weight parameter $\lambda$ in the multi-task loss and the ratio $p$ of the ordered videos in a batch. These experiments are conducted under the P$\to$N-CV setting.
Fig. \ref{fig:hyper} (a) and (b) studies the impact of $\lambda_t$ and $p_t$, the hyper-parameters in the temporal Cubism approach.
In Fig. \ref{fig:hyper} (a) is shown the change of the recognition accuracy when $p_t$ is fixed to 0.8 while $\lambda_t$ varies from 0 to 10. It is noticed that the best performance is acquired when $\lambda_t = 0.1$.
The value of $\lambda_t$ that is less than 0.1 degrades the effect of the loss of the self-supervised task and furthermore the effect of the self-supervision based method. But the performance is still ameliorated based on the baseline thanks to the data augmentation. On the other hand, a greater value of $\lambda_t$ will possibly make the auxiliary task overwhelm the main task, which will also result in the decrement of the performance. As evidence, the performance plummets sharply when $\lambda = 10$.
Moreover, we studied the impact of $p_t$ and show in Fig. \ref{fig:hyper} (b) the variation of the obtained accuracy in the case that $\lambda_t$ is fixed to 0.1 while $p_t$ varies from 0.2 to 1. It is found that various scales of $p_t$ result in consistent boosting of the adaptation performance base on the baseline while the best performance is achieved with $p_t = 0.8$.
Meanwhile, we notice that the performance is slightly poorer for the relatively greater value of $p_t$.
This may be explained by that greater $p_t$ degrades the role of the auxiliary task and the method will fall back to the baseline when $p_t = 1$.
Likewise, we verified the impact of the hyper-parameters in the spatial Cubism approach, $\lambda_s$ and $p_s$, and received the similar curves shown in Fig. \ref{fig:hyper} (c) and (d). The most effective parameters for spatial Cubism approach appear to be $\lambda_s = 0.1$ and $p_s = 0.6$.

\begin{figure}[t]
\includegraphics[width = \linewidth]{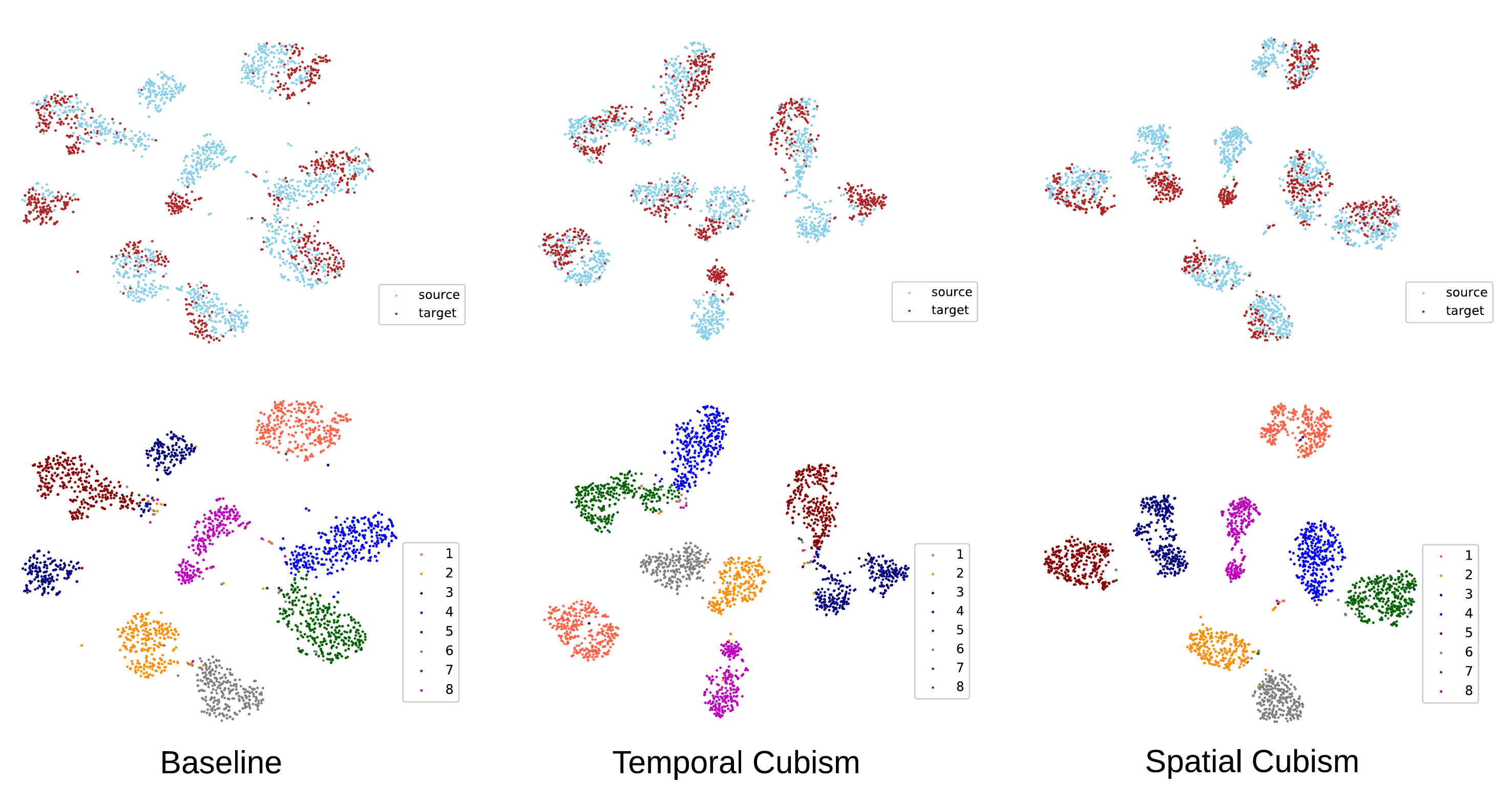} 
\caption{The t-SNE visualization of the feature distribution of the samples from the source and target domains. The samples are extracted from eight action categories.
The plots on the left color the samples in the source domain with blue and the samples in the target domain with brown,
while the plots on the right use different colors to represent the points from the various action classes.
}
\label{fig:tsne}
\vspace{-0.15 cm}
\end{figure}

\noindent \textbf{Analysis of the segment number:}
In section \ref{subsection:tem_cub}, a video is divided into $N$ ($N=3$) segments uniformly in the temporal domain to perform temporal Cubism.
Here we evaluate other numbers for $N$ ($N=2$ and $N=4$) under P$\to$N-CV setting and present the results in Table~\ref{tab:N_seg}.
We can observe that $N=3$ outperforms other segment numbers.
This can be attributed to dividing a video into three segments making the amount of permutation categories properly, which is vital to make the auxiliary task neither too simple nor too complicated.

\begin{table}[t]
    \centering
    \small
    \setlength{\tabcolsep}{7mm}
    \caption{The experimental results for dividing a video into $N$ segments under P$\to$N-CV setting.}
	\begin{tabular}{ c|c|c|c}
	\hline
	Segment number $N$ & 2 & 3 & 4\\
	\hline
	Accuracy & 56.9 & \textbf{58.3} & 56.5 \\
	\hline
    \end{tabular}
    \vspace{-0.1cm}
	\label{tab:N_seg}  
\end{table}

\noindent \textbf{t-SNE Visualization:}
We visualize the distribution of both domains extracted from some experiments under the P$\to$N-CV setting using t-SNE\cite{MaatenLaurens_van_der2008JMLR_t-SNE} in Fig. \ref{fig:tsne}. 
It is shown that the samples from different domains are not that well-aligned through the baseline model though the source samples could be finely grouped into clusters, while the distributions of two domains get much closer when applied temporal or spatial Cubism methods. These plots intuitively demonstrate the effectiveness of our methods.

\begin{figure*}[t]
\includegraphics[width = \linewidth]{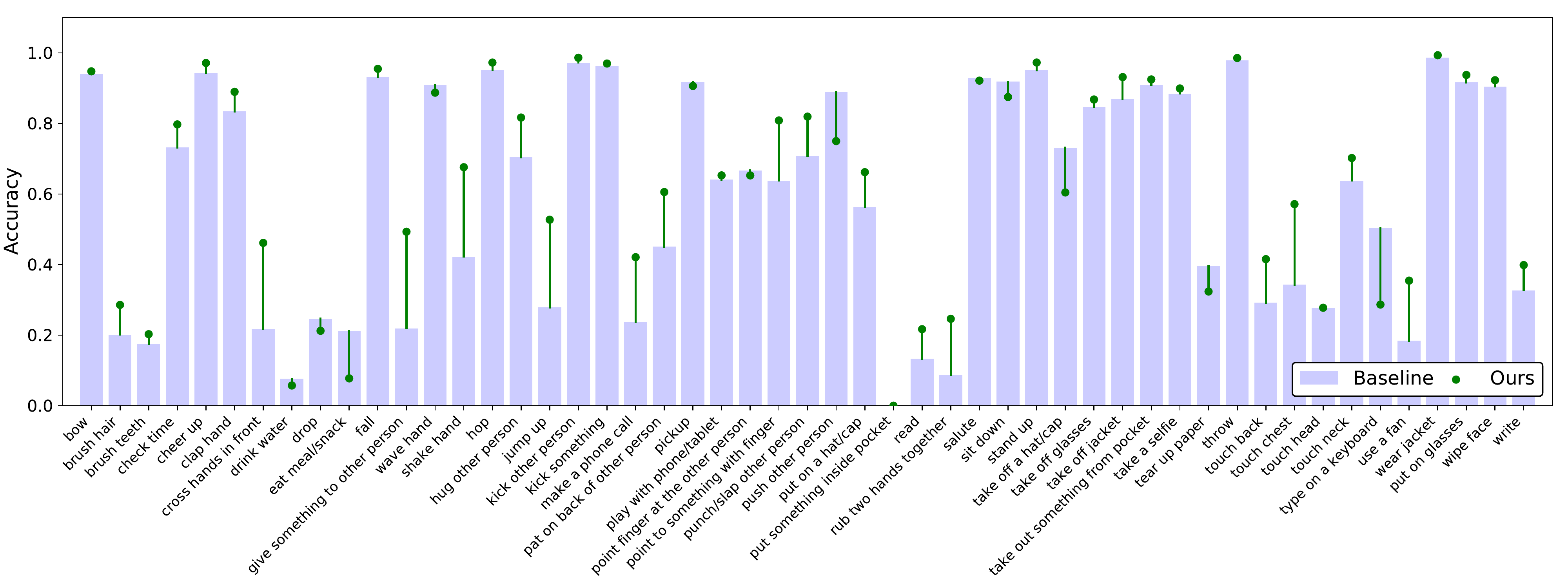}
\caption{The action recognition accuracy on the individual actions under UDA setting from NTU RGB+D to PKU-MMD (CV). The action label can be best viewed by zooming the PDF file.}
\label{fig:actions}  
\end{figure*}

\noindent \textbf{Analysis of Different Actions:}
Fig. \ref{fig:actions} illustrates the performance of TS-Cub on the individual action categories compared to the baseline.
Our method achieves consistent improvement on most of the action categories and outperforms the best on the actions \textit{cross hand in front}, \textit{give something to other person}, \textit{shake hand}, \textit{jump up} and \textit{touch chest}. On these actions, the baseline achieves relatively poorer performance at first and our method manages to enhance the adaptation performance.
On the other side, our TS-Cub fails to reach the baseline on the actions \textit{eat meal/snack}, \textit{push other person}, \textit{take off hat/cap} and \textit{type on a keyboard}. Actually, the videos from \textit{eat meal/snack} have an unduly longer duration and the videos from \textit{push other person} have a shorter duration than the videos from other categories, which may bring extra difficulty for the temporal stream. Action \textit{take off hat/cap} and \textit{type on a keyboard} hold a little more tiny spatial features that may confuse the spatial stream.
For the baseline and our proposed TS-Cub approach, we find that they both fail to recognize the action
\textit{put something inside pocket}. This is because in the NTU dataset, this action involves two people (one person puts something inside the pocket of another person). However, in the PKU dataset this is a single person action (one person puts something inside the pocket of himself/herself). This failure case suggests the limitation of using skeleton-based data as input for recognizing action involving semantic object (\textit{e.g.,} interaction with the pocket). This issue would be tackled by further leveraging the RGB modality. We will explore this interesting direction in the future.

\begin{table}[t]
	\centering
	\caption{Comparison of the skeleton-based action recognition accuracy (\%) of single action and interacted action on the N$\to$P-CV setting.}
	\setlength{\tabcolsep}{2.3pt}
	\vspace{-0.15 cm}
	\begin{tabular}{l|c|c|c}
		\hline
		Method & single action & interacted action &	overall\\
		\hline
		Baseline (source only) & 59.1 & 66.1 & 59.6 \\
		\hline		
		Tem-Cub (Ours)        & 64.7 & 72.0 & 65.2\\
		Spa-Cub (Ours)        & 61.2 & 71.8 & 61.2\\
		TS-Cub (Ours)         & 64.8 & 73.0 & 65.5\\
		\hline
	\end{tabular}
	\vspace{-0.2cm}
	\label{tab:single_multiple}
\end{table}

\noindent \textbf{Analysis of the Number of People:}
In NTU and PKU datasets, there are numbers of interacted actions that involves multiple people in each video (\textit{e.g.,} \textit{shaking hands}, \textit{hugging}, \textit{etc}). 
For these actions, we follow~\cite{DBLP:conf/ijcai/LiZXP18} to apply element-wise maximum operation to fuse the features of multiple people.
Furthermore,
we compare the experimental results under the N$\to$P-CV setting of the single action and the interacted action in Table~\ref{tab:single_multiple}. 
We observe that our method obtains larger improvements over the interacted action (66.1\% $\to$ 73.0\%) than those over the single action (59.1\% $\to$ 64.8\%).
These experimental results demonstrate the generalized capability of our TS-Cub model, which can effectively deal with the multiple people cases with the element-wise maximum operation.

\begin{table}[t]
	\centering
	\caption{Comparison of the skeleton-based action recognition accuracy (\%) when using different number of joints $N_j$. Experiments are conducted under the N$\to$P-CV setting.}
	\setlength{\tabcolsep}{7pt}
	\vspace{-0.15 cm}
	\begin{tabular}{l|c|c|c|c}
		\hline
		Method & $N_j$ = 25 & $N_j$ = 22 & $N_j$ = 18 & $N_j$ = 12\\
		\hline
		Tem-Cub (Ours)         & 65.2 & 65.8 & 64.5 & 63.5 \\
		Spa-Cub (Ours)         & 61.2 & 61.9 & 61.8 & 60.2 \\
		TS-Cub (Ours)          & 65.5 & 65.4 & 64.7 & 63.3 \\
		\hline
	\end{tabular}
	\label{tab:number_of_joint}
\end{table}

\noindent \textbf{Analysis of the Number of Joints:}
We further conduct experiments to ablate the number of joints $N_j$ in skeleton-based video\footnote{Based on the 25 joints used in~\cite{DBLP:conf/cvpr/ShahroudyLNW16}, we remove the joints ``middle of the spine'', ``left thumb'' and ``right thumb'' for $N_j$ = 22, remove the joints ``middle of the spine'', ``left hand'', ``right hand'', ``left ankle'', ``right ankle'', ``left thumb'' and ``right thumb'' for $N_j$ = 18, remove the joints ``middle of the spine'', ``left hand'', ``right hand'', ``left ankle'', ``right ankle'', ``left wrist'', ``right wrist'', ``left elbow'', ``right elbow'', ``left knee'', ``right knee'', ``left thumb'' and ``right thumb'' for $N_j$ = 12.}.
As shown in Table~\ref{tab:number_of_joint}, our final TS-Cub model achieves better results with more joints as input. When only using 12 major joints, it can also obtain a comparable performance with the result of $N_j$ = 25, indicating its robustness to the number of joints. We also find that using less joints could achieve better results in some cases (\textit{e.g.,} $N_j=22$ versus $N_j=25$), this is because the absent joints (\textit{e.g.,} left thumb and right thumb) sometimes would be redundant or even bring noise to the final action results.

\begin{table}[t]
	\centering
	\caption{Comparison of the skeleton-based action recognition accuracy (\%) when influenced by Gaussian noise with different standard deviations $\sigma$. Experiments are conducted under the N$\to$P-CV setting.}
	\setlength{\tabcolsep}{5pt}
	\begin{tabular}{l|c|c|c|c}
		\hline
		Method & $\sigma=0$ & $\sigma=0.001$ & $\sigma=0.01$ & $\sigma=0.1$
		\\
		\hline		
		Tem-Cub (Ours) & 65.2 & 63.3 & 62.0 & 43.3 \\
		Spa-Cub (Ours) & 61.2 & 60.8 & 60.0 & 39.9 \\
		TS-Cub (Ours)  & 65.5 & 63.1 & 62.5 & 42.5 \\
		\hline
	\end{tabular}
	\label{tab:gaussian}
\end{table}

\noindent \textbf{Analysis of the Gaussian Noise:}
To evaluate the robustness of our method, we add Gaussian noise to the input video. Specifically, we first normalize the input data into the scale of [-1, 1], add Gaussian noise with zero mean and different standard deviations $\sigma$ to them, and re-scale data to the original interval. As shown in Table \ref{tab:gaussian}, with perturbation of $\sigma=0.001$ and $\sigma=0.01$, our algorithm achieves comparable performance with that of $\sigma=0$. 
But with more noise, TS-Cub has a noticeable decrease from 65.5\% ($\sigma=0$) to 42.5\% ($\sigma=0.1$).

\begin{table*}[!t]
	\centering
	\caption{Exploration on other self-supervised learning tasks.
Source Domain: PKU-MMD (P) dataset. Target Domain: NTU-RGB+D (N51) dataset (CV setting).
For the definition of the tasks, refer to section \ref{subsection:expe_res_and_anal} for details.}
	\begin{tabular}{l|ccc||cc}
		\hline
		Task  &	Spa-Cub &	Spa-Jigsaw &	Freezing Game &	Tem-Cub &	Tem-Flip \\
		\hline
		Accuracy &	56.5 &		54.2 &		    55.3      &	58.3 &	57.1  \\
		\hline
		Fusion way & WRSM & WGM   & MP  & TS-Cub (WAM) &	Coupled-Cub \\
		\hline
		Accuracy & 51.9 & 58.9 & 58.5 & \textbf{59.2} &	55.6 \\
		\hline
	\end{tabular}
	\label{tab:other_self_super_tasks}
\end{table*}

\noindent \textbf{Exploration of Other Self-supervised Learning Tasks:}
Besides our proposed Cubism tasks, we also explore other self-supervised learning tasks but receive not that satisfactory results. The comparison of results adopting different supervised tasks is shown in Table \ref{tab:other_self_super_tasks}.
For instance, we consider the task \textit{Tem-Flip} to distinguish the ordered videos from the temporally inverse videos. However, it is hard to discriminate between such actions like ordered \textit{put something inside pocket} and inverse \textit{take out something from pocket}. Hence this task cannot be applied to all action categories and fails to lead to higher performance.
We explore a task named as \textit{Spa-Jigsaw} in the temporal domain.
There are a number of joints comprising a body, which are stored in a linear list in practice. Hence we simply uniformly divide that list into 3 segments and permute them. This way of permutation thoroughly breaks the spatial relation of the joints and thereby achieves a slightly poor result. 
Meanwhile, we try another way called \textit{Freezing Game} to build augmented data by freezing the pose of the arms and the legs from the first frame during the whole video. However, as several actions do not comprise large-amplitude motions at first, this task seems to be so difficult that the importance of the expected classification task gets degraded.
Though spatial rotation is a conventional transformation, we do not exploit it further as we have taken it as a part of the data pre-processing.
Additionally, several approaches are investigated to combine Tem-Cub and Spa-Cub. Firstly, we explore some softmax scores fusion approaches like $\lambda_{1}\textbf{s}_{1}+\lambda_{2}\textbf{s}_{2}$, $\sqrt{\lambda_{1}{\textbf{s}_{1}}^{2}+\lambda_{2}{\textbf{s}_{2}}^{2}}$, ${\textbf{s}_{1}}^{\lambda_{1}}{\textbf{s}_{2}}^{\lambda_{2}}$, and max-pool$(\textbf{s}_{1},\textbf{s}_{2})$, named as Weighted Arithmetic Mean (WAM), Weighted Root Squared Mean (WRSM), Weighted Geometric Mean (WGM) and Max Pooling (MP) orderly in Table \ref{tab:other_self_super_tasks}. Here $\textbf{s}_{1}$ and $\textbf{s}_{2}$ denote temporal and spatial softmax scores, and $\lambda_{1}=0.6, \lambda_{2}=0.4$ are two hyper-parameters. We find that simply add temporal and spatial softmax scores achieves the best result and name it \textit{TS-Cub}. Besides, there is another method that applies the temporal and spatial transformations simultaneously to the training samples. This combination way couples the temporal Cubism and spatial Cubism and is named as \textit{Coupled-Cub}. \textit{Coupled-Cub} will considerably increase the number of the permutation labels and produce more disordered samples, which raises the difficulty of the auxiliary task as well.
As a conclusion of our exploration, the auxiliary task is not supposed to be too simple or too difficult. A too simple task has not got enough ability to draw the two domains close while a too difficult task could overwhelm the original classification task.
In other words, an inadequate self-supervised task could result in an even worse adaptation performance thus choosing an appropriate additional task is crucial for such a self-supervision based approach.

\begin{table*}[t]
	\centering
	\small
	\caption{A new cross-dataset skeleton-based action recognition setting where the data from target domain are totally unavailable during training.}
	\setlength{\tabcolsep}{7pt}
	\begin{tabular}{l | c c |c c  }
	\hline
	 & \multicolumn{2}{c|}{Training} &  \multicolumn{2}{c}{Testing}\\
	 & Source& Clips  & Target & Clips\\ 
	 \hline
	P$\to$N-CV & P & 21544   & N51-CV-test& 16092\\
	P$\to$N-CS & P &21544  & N51-CS-test& 14013\\
	N$\to$P-CV &  N51 & 48081  & P-CV-test&7188 \\
	N$\to$P-CS &  N51 & 48081  & P-CS-test&2704 \\
	\hline
	N$\to$K & N12 & 11256  & K-test & 787 \\
	K$\to$N & K & 9699  & N12-CV-test & 3780 \\
	\hline
	\end{tabular}
	\label{tab:new_setting}  
	\vspace{-0.1cm}
\end{table*}

\begin{table*}[t]
	\small
	\centering
	\caption{The experimental results for training without target domain data.} 
	\setlength{\tabcolsep}{3pt}
	\vspace{-0.2cm}
	\begin{tabular}{l|cccc|cc}
		\hline
		Method &	P$\to$N-CV & P$\to$N-CS &	N$\to$P-CV& N$\to$P-CS &  N$\to$K  & K$\to$N \\
		\hline
		Target Only & 91.3 & 84.6 &	95.7 & 93.1 & 40.8 & 89.1 \\
		Source Only & 54.9 & 50.5 &	59.6 & 57.6 & 14.4 & 22.5 \\
		\hline
		Tem-Cub &	57.1&52.7 &	62.7&60.0 & \textbf{15.6} & 25.5 \\
		Spa-Cub  &	54.7&50.8 & 61.9&59.6 & 15.0 & 25.1 \\
		TS-Cub  &	\textbf{57.7}&\textbf{53.8} &	\textbf{63.4}&\textbf{61.3} & 15.5& \textbf{25.6} \\
		\hline
	\end{tabular}
	\label{tab:exp_res}
	\vspace{-0.35cm}
\end{table*}

\subsection{Training the TS-Cub without Target Domain Data}
\label{without_target_domain}
 We make further exploration about testing our proposed TS-Cub under a more challenging cross-dataset setting, which assumes that data from target domain are totally unavailable during the training period. We detail this unsupervised domain adaptation setting in Table~\ref{tab:new_setting}. During the training phase, the permuted video samples from source domain are delivered into the network along with the ordering data. The final losses are composed of the main classification task and the auxiliary Cubism task.

We present the experimental results in Table \ref{tab:exp_res}, other compared methods in the previous section are absent because they all required the target data during training, which is not available in this setting.
As shown in Table \ref{tab:exp_res}, our TS-Cub consistently outperforms the baseline \textit{Source Only} on the six tasks, which indicates its robustness for cross-dataset skeleton-based action recognition.

\begin{table}[t]
	\centering
	\caption{Combining the Cubism with CDAN~\cite{CDAN} under the UDA setting.}
	\setlength{\tabcolsep}{1.0pt}
	\begin{tabular}{l|cc|cc|cc}
		\hline
		Method &	P$\to$N-CV & P$\to$N-CS &	N$\to$P-CV& N$\to$P-CS & N$\to$K & K$\to$N \\
		\hline
		Tem-Cub (Ours) &	58.3& 52.5 & 65.2 & 62.8 & 16.4 & 29.3 \\
		Spa-Cub (Ours) &	56.5& 51.0 & 61.2 & 59.3 & 14.6 & 22.9 \\
		TS-Cub (Ours) &		59.2 & 53.6 & 65.5& 63.3 & \textbf{16.8} & \textbf{29.6}\\
		\hline
		Tem-Cub (Ours) + CDAN~\cite{CDAN,DBLP:conf/eccv/ChoiSSH20} & \textbf{60.2} & 56.5 & 67.6 & 64.6 & 16.2 & 21.5 \\
		Spa-Cub (Ours) + CDAN~\cite{CDAN} & 59.3 & 56.4 & 68.3 & 65.4 & 15.6 & 26.8 \\	
		TS-Cub (Ours) + CDAN~\cite{CDAN} & 59.5 & \textbf{57.8} & \textbf{68.9} & \textbf{66.1} & 16.6 & 27.1 \\
		\hline
	\end{tabular}
	\vspace{-0.2cm}
	\label{tab:cubism_and_cdan}
\end{table}

\subsection{Combining the Cubism with Adversarial Learning Method}
Recently, Choi \textit{et al.}~\cite{DBLP:conf/eccv/ChoiSSH20} study the cross-dataset action recognition problem by combining the domain adversarial task with the clip order prediction task.
Motivated by this work, we further conduct experiments to see whether our self-supervised pretext tasks are complementary with the conventional domain adversarial task. 
Since~\cite{DBLP:conf/eccv/ChoiSSH20} is designed for RGB video while our work focus on skeleton-based video, we use HCN~\cite{DBLP:conf/ijcai/LiZXP18} as the backbone similar with TA$^3$N~\cite{video_DA}.
Then we perform temporal Cubism at raw data level as well as apply adversarial learning in feature level based on CDAN~\cite{CDAN} (denoted as ``Tem-Cub (Ours) + CDAN'' in Table \ref{tab:cubism_and_cdan}).
We also conduct experiment on combining our Spa-Cub with CDAN (\textit{i.e.,} ``Tem-Cub (Ours) + CDAN''), and ensembling the results of ``Tem-Cub (Ours) + CDAN'' and ``Spa-Cub (Ours) + CDAN'' (\textit{i.e.,} ``TS-Cub (Ours) + CDAN'').

We present the compared results in Table ~\ref{tab:cubism_and_cdan}. 
On P$\leftrightarrow$N setting, we find the performance could be further improved by combining our approach with CDAN~\cite{CDAN}, which shows the complementary characteristics of our method and adversarial approach.
However, on the N$\to$K setting, we found that the performance drops slightly when combining with CDAN. This might attribute to the videos from the Kinetics are collected in the wild, and the skeleton-based inputs are obtained from 2D pose estimation algorithm rather than 3D sensor for NTU and PKU datasets. In this case, the adversarial approach (\textit{e.g.,} CDAN) might have more difficulty in dealing with this kind of data with more noise. In comparison, our method is more robust to generalize to this more challenging scenario.

%---------------------------------------------------------
\section{Conclusions}

In this paper, we have investigated the unsupervised domain adaptation setting for skeleton-based action recognition.
In order to reduce the domain shift between different datasets, we have devised a self-supervised learning approach based on temporal spatial Cubism.
Both quantitative and qualitative experimental results have demonstrated the effectiveness of our method.
We expect this work to provide a new direction for skeleton-based action recognition, and inspire applications to other related tasks, such as group activity recognition~\cite{DBLP:journals/tip/TangLWYZ19}, action quality assessment~\cite{DBLP:conf/cvpr/TangNZZLWZ20} and instructional video analysis~\cite{coin}.

\bibliographystyle{ACM-Reference-Format}
\bibliography{sample-base}

\end{document}